\documentclass[sigconf]{acmart}
\renewcommand\footnotetextcopyrightpermission[1]{} 

\usepackage{booktabs} 
\usepackage{url}
\usepackage{amsmath}
\usepackage{amssymb}
\usepackage{graphicx}
\usepackage{hyperref}
\usepackage{wrapfig}
\renewcommand{\vec}[1]{\mathbf{#1}}
\usepackage{cleveref}
\usepackage[bottom]{footmisc}

\usepackage{listings}
\usepackage{color}

\definecolor{dkgreen}{rgb}{0,0.6,0}
\definecolor{gray}{rgb}{0.5,0.5,0.5}
\definecolor{mauve}{rgb}{0.58,0,0.82}

\setcopyright{none}

\acmConference[]{}{}{}
\acmYear{2017}
\copyrightyear{2017}

\begin{document}
\title{Towards Building an Intelligent Anti-Malware System: A Deep Learning Approach using Support Vector Machine (SVM) for Malware Classification}

\author{Abien Fred M. Agarap}
\email{abienfred.agarap@gmail.com}

\begin{abstract}
Effective and efficient mitigation of malware is a long-time endeavor in the information security community. The development of an anti-malware system that can counteract an unknown malware is a prolific activity that may benefit several sectors. We envision an intelligent anti-malware system that utilizes the power of deep learning (DL) models. Using such models would enable the detection of newly-released malware through mathematical generalization. That is, finding the relationship between a given malware $\vec{x}$ and its corresponding malware family $y$, $f: \vec{x} \mapsto y$. To accomplish this feat, we used the Malimg dataset\cite{nataraj2011malware} which consists of malware images that were processed from malware binaries, and then we trained the following DL models\footnote{Code available at https://github.com/AFAgarap/malware-classification} to classify each malware family: CNN-SVM\cite{tang2013deep}, GRU-SVM\cite{agarap2017neural}, and MLP-SVM. Empirical evidence has shown that the GRU-SVM stands out among the DL models with a predictive accuracy of $\approx$84.92\%. This stands to reason for the mentioned model had the relatively most sophisticated architecture design among the presented models. The exploration of an even more optimal DL-SVM model is the next stage towards the engineering of an intelligent anti-malware system.
\end{abstract}



\maketitle

\section{Introduction}
Effective and efficient mitigation of malware is a long-time endeavor in the information security community. The development of an anti-malware system that can counteract an unknown malware is a prolific activity that may benefit several sectors.\\
\indent	To intercept an unknown malware or even just an unknown variant is a laborious task to undertake, and may only be accomplished by constantly updating the anti-malware signature database. The mentioned database contains the information on all known malware by the particular system\cite{shelly2011discovering}, which is then used for malware detection. Consequently, newly-released malware which are not yet included in the database will go undetected.\\
\indent	We envision an intelligent anti-malware system that employs a deep learning (DL) approach which would enable the detection of newly-released malware through its capability to generalize on data. Furthermore, we amend the conventional DL models to use the support vector machine (SVM) as their classification function.
\indent	We take advantage of the Malimg dataset\cite{nataraj2011malware} which consists of visualized malware binaries, and use it to train the DL-SVM models to classify each malware family.

\section{Methodology}

\subsection{Machine Intelligence Library}
Google TensorFlow\cite{tensorflow2015-whitepaper} was used to implement the deep learning algorithms in this study, with the aid of other scientific computing libraries: matplotlib\cite{Hunter:2007}, numpy\cite{walt2011numpy}, and scikit-learn\cite{scikit-learn}. 

\subsection{The Dataset}

The deep learning (DL) models in this study were evaluated on the Malimg dataset\cite{nataraj2011malware}, which consists of 9,339 malware samples from 25 different malware families. Table \ref{table: malimg-dataset} shows the frequency distribution of malware families and their variants in the Malimg dataset\cite{nataraj2011malware}.
\begin{figure}[!htb]
	\minipage{0.45\textwidth}
	\centering
		\includegraphics[width=\linewidth]{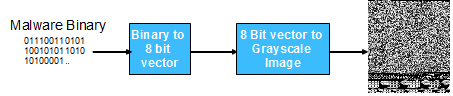}
		\caption{Image from \cite{nataraj2011malware}. Visualizing malware as a grayscale image.}
		\label{malimg}
	\endminipage\hfill
\end{figure}

\begin{table}[!htb]
\centering
\caption{Malware families found in the Malimg Dataset\cite{nataraj2011malware}.}
		\begin{tabular}{cccc}
		\toprule
		No. & Family & Family Name & No. of Variants \\
		\midrule
		01	&	Dialer				&	Adialer.C 		&	122 \\
		02	&	Backdoor			&	Agent.FYI 		&	116 \\
		03	&	Worm	 			&	Allaple.A 		&	2949 \\
		04	&	Worm 				&	Allaple.L 		&	1591 \\
		05	&	Trojan 				&	Alueron.gen!J 	&	198 \\
		06	&	Worm:AutoIT 		&	Autorun.K 		&	106 \\
		07	&	Trojan 				&	C2Lop.P 		&	146 \\
		08	&	Trojan 				&	C2Lop.gen!G 	&	200 \\
		09	&	Dialer 				&	Dialplatform.B 	&	177 \\
		10	&	Trojan Downloader	&	Dontovo.A 		&	162 \\
		11	&	Rogue 				&	Fakerean 		&	381 \\
		12	&	Dialer				&	Instantaccess 	&	431 \\
		13	&	PWS 				&	Lolyda.AA 1 	&	213 \\
		14	&	PWS 				&	Lolyda.AA 2 	&	184 \\
		15	&	PWS 				&	Lolyda.AA 3 	&	123 \\
		16	&	PWS					&	Lolyda.AT 		&	159 \\
		17	&	Trojan				&	Malex.gen!J 	&	136 \\
		18	&	Trojan Downloader	&	Obfuscator.AD 	&	142 \\
		19	&	Backdoor			&	Rbot!gen 		&	158 \\
		20	&	Trojan				&	Skintrim.N 		&	80 \\
		21	&	Trojan Downloader	&	Swizzor.gen!E 	&	128 \\
		22	&	Trojan Downloader	&	Swizzor.gen!I 	&	132 \\
		23	&	Worm				&	VB.AT 			&	408 \\
		24	&	Trojan Downloader	&	Wintrim.BX 		&	97 \\
		25	&	Worm				&	Yuner.A 		&	800 \\
		\bottomrule
		\end{tabular}\\
		\label{table: malimg-dataset}
\end{table}

Nataraj et al. (2011)\cite{nataraj2011malware} created the Malimg dataset by reading malware binaries into an 8-bit unsigned integer composing a matrix $M \in \mathbb{R}^{m \times n}$. The said matrix may be visualized as a grayscale image having values in the range of $[0, 255]$, with 0 representing \textit{black} and 1 representing \textit{white}.

\subsection{Dataset Preprocessing}
Similar to what \cite{garcia2016random} did, the malware images were resized to a 2-dimensional matrix of $32 \times 32$, and were flattened into a $n \times n$-size array, resulting to a $1 \times 1024$-size array. Each feature array was then labelled with its corresponding indexed malware family name (i.e. $0-24$). Then, the features were standardized using Eq. \ref{student-t-stat}.
\begin{equation}\label{student-t-stat}
z=\dfrac{X - \mu}{\sigma}
\end{equation}

where $X$ is the feature to be standardized, $\mu$ is its mean value, and $\sigma$ is its standard deviation. The standardization was implemented using \texttt{StandardScaler().fit\_transform()} of \texttt{scikit-learn}\cite{scikit-learn}. Granted that the dataset consists of images, and standardization may not be suitable for such data, but take note that the images originate from malware binary files. Hence, the features are not technically images to begin with.

\subsection{Computational Models}\label{computational-models}
This section presents the deep learning (DL) models, and the support vector machine (SVM) classifier used in the study.

\subsubsection{Support Vector Machine (SVM)}\label{svm}
The support vector machine (SVM) was developed by Vapnik\cite{Cortes} for binary classification. Its objective is to find the optimal hyperplane $f(\vec{w}, \vec{x}) = \vec{w} \cdot \vec{x} + b$ to separate two classes in a given dataset, with features $\vec{x} \in \mathbb{R}^{m}$.

SVM learns the parameters $\vec{w}$ and $b$ by solving the following constrained optimization problem:
\begin{equation} \label{constrained-l1}
min \dfrac{1}{p}\vec{w}^{T}\vec{w} + C \sum_{i = 1}^{p} \xi_i
\end{equation}
\begin{align}
s.t\ y_{i}'(\vec{w} \cdot \vec{x} + b) \geq 1 - \xi_i\\
\xi_i \geq 0, i = 1, \dots, p
\end{align}

where $\vec{w}^{T} \vec{w}$ is the Manhattan norm (also known as L1 norm), $C$ is the penalty parameter (may be an arbitrary value or a selected value using hyper-parameter tuning), and $\xi$ is a cost function.

The corresponding unconstrained optimization problem of Eq. \ref{constrained-l1} is given by Eq. \ref{l1-svm}.
\begin{equation} \label{l1-svm}
min \dfrac{1}{p}\vec{w}^{T}\vec{w} + C \sum_{i = 1}^{p} max\big(0, 1 - y_{i}'(\vec{w}^{T}\vec{x}_{i}+b)\big)
\end{equation}

\indent where $\vec{y'}$ is the actual label, and $\vec{w}^T\vec{x} + b$ is the predictor function. This equation is known as L1-SVM, with the standard hinge loss. Its differentiable counterpart, L2-SVM (given by Eq. \ref{l2-svm}), provides more stable results\cite{tang2013deep}.

\begin{equation}\label{l2-svm}
min \dfrac{1}{p}\|\vec{w}\|_{2}^{2} + C \sum_{i = 1}^{p} max\big(0, 1 - y_{i}'(\vec{w}^{T}\vec{x}_{i}+b)\big)^{2}
\end{equation}

where $\|\vec{w}\|_{2}$ is the Euclidean norm (also known as L2 norm), with the squared hinge loss.

Despite intended for binary classification, SVM may be used for multinomial classification as well. One approach to achieve this is the use of kernel tricks, which converts a linear model to a non-linear model by applying kernel functions such as radial basis function (RBF). However, for this study, we utilized the linear L2-SVM for the multinomial classification problem. We then employed the \textit{one-versus-all} (OvA) approach, which treats a given class $c_{i}$ as the positive class, and others as negative class.\\
\indent	Take for example the following classes: \textit{airplane, boat, car}. If a given image belongs to the \textit{airplane} class, it is taken as the positive class, which leaves the other two classes the negative class.\\
\indent	With the OvA approach, the L2-SVM serves as the classifier of each deep learning model in this study (CNN, GRU, and MLP). That is, the learning parameters \textit{weight} and \textit{bias} of each model is learned by the SVM.

\subsubsection{Convolutional Neural Network}\label{cnn}
Convolutional Neural Networks (CNNs) are similar to feedforward neural networks for they also consist of hidden layers of neurons with ``learnable'' parameters. These neurons receive inputs, performs a dot product, and then follows it with a non-linearity such as $sigmoid$ or $tanh$. The whole network expresses the mapping between raw image pixels $\vec{x} \in \mathbb{R}^{m}$ and class scores $y$, $f: \vec{x} \mapsto y$. For this study, the CNN architecture used resembles the one laid down in \cite{tensorflow_2017}:

\begin{enumerate}
\item INPUT: $32 \times 32 \times 1$
\item {\color{green}CONV5: $5 \times 5$ size, 36 filters, 1 stride}
\item {\color{red}LeakyReLU: $max(0.01h_{\theta}(x)), h_{\theta}(x))$}
\item {\color{blue}POOL: $2 \times 2$ size, 1 stride}
\item {\color{green}CONV5: $5 \times 5$ size, 72 filters, 1 stride}
\item {\color{red}LeakyReLU: $max(0.01h_{\theta}(x)), h_{\theta}(x))$}
\item {\color{blue}POOL: $2 \times 2$ size, 1 stride}
\item {\color{orange}FC: 1024 Hidden Neurons}
\item {\color{red}LeakyReLU: $max(0.01h_{\theta}(x)), h_{\theta}(x))$}
\item {\color{purple}DROPOUT: $p = 0.85$}
\item {\color{orange}FC: 25 Output Classes}
\end{enumerate}

The modification introduced in the architecture design was the size of layer inputs and outputs (e.g. input of $32 \times 32 \times 1$ instead of $28 \times 28 \times 1$, and output of 25 classes), the use of \texttt{LeakyReLU} instead of \texttt{ReLU}, and of course, the introduction of L2-SVM as the network classifier instead of the conventional Softmax function. This paradigm of combining CNN and SVM was actually proposed by Tang (2013)\cite{tang2013deep}.

\subsubsection{Gated Recurrent Unit}\label{gru}
Agarap (2017)\cite{agarap2017neural} proposed a neural network architecture combining the gated recurrent unit (GRU)\cite{Cho} variant of a recurrent neural network (RNN) and the support vector machine (SVM)\cite{Cortes} for the purpose of binary classification.
\begin{equation}\label{z-gate}
z	=	\sigma(\vec{W}_{z} \cdot [h_{t - 1}, x_{t}])
\end{equation}
\begin{equation}\label{r-gate}
r	=	\sigma(\vec{W}_{r} \cdot [h_{t - 1}, x_{t}])
\end{equation}
\begin{equation}\label{candidate-value}
\tilde{h}_{t}	=	tanh(\vec{W} \cdot [r_{t} * h_{t - 1}, x_{t}])
\end{equation}
\begin{equation}\label{new-value}
h_{t}	=	(1 - z_{t}) * h_{t - 1} + z_{t} * \tilde{h}_{t}
\end{equation}

where $z$ and $r$ are the \textit{update gate} and \textit{reset gate} of a GRU-RNN respectively, $\tilde{h}_{t}$ is the candidate value, and $h_{t}$ is the new RNN cell state value\cite{Cho}. In turn, the $h_{t}$ is used as the predictor variable $x$ in the L2-SVM predictor function (given by $\vec{w}\vec{x} + b$) of the network instead of the conventional Softmax classifier.

\subsubsection{Multilayer Perceptron}\label{mlp}
The perceptron model was developed by Rosenblatt (1958)\cite{rosenblatt1958perceptron} based on the neuron model by McCulloch \& Pitts (1943)\cite{mcculloch1943logical}. A perceptron may be represented by a linear function (given by Eq. \ref{mlp-linear}), which is then passed to an activation function such as \textit{sigmoid} $\sigma$, \textit{sign}, or \textit{tanh}. These activation functions introduce non-linearity (except for the \textit{sign} function) to represent complex functions.\\
\indent	As the term itself implies, a multilayer perceptron (MLP) is a neural network that consists of hidden layers of perceptrons. In this study, the activation function used was the \texttt{LeakyReLU}\cite{maas2013rectifier} function (given by Eq. \ref{leaky-relu}).
\begin{align}\label{mlp-linear}
h_{\theta}(x)	&=	\sum_{i = 0}^{n} \vec{\theta}_{i} \vec{x}_{i} + b
\end{align}
\begin{align}\label{leaky-relu}
f\big(h_{\theta}(x)\big)	&=	max\big(0.01h_{\theta}(x), h_{\theta}(x)\big)
\end{align}

\indent The learning parameters \texttt{weight} and \texttt{bias} for each DL model were learned by the L2-SVM using the loss function given by Eq. \ref{l2-svm}. The computed loss is then minimized through Adam\cite{Kingma} optimization. Then, the decision function $f(x) = sign(\vec{w}\vec{x} + b)$ produces a vector of scores for each malware family. In order to get the predicted labels $y$ for a given data $x$, the $argmax$ function is used (see Eq. \ref{argmax}).
\begin{align}\label{argmax}
\vec{y'}	&=	argmax\big(sign(\vec{w}\vec{x} + b)\big)
\end{align}

\indent	The $argmax$ function shall return the indices of the highest scores across the vector of predicted classes $\vec{w}\vec{x} + b$.

\subsection{Data Analysis}
There were two phases of experiment for this study: (1) training phase, and (2) test phase. All the deep learning algorithms described in Section \ref{computational-models} were trained and tested on the Malimg dataset\cite{nataraj2011malware}. The dataset was partitioned in the following fashion: 70\% for training phase, and 30\% for testing phase.\\
\indent	The variables considered in the experiments were the following: 
\begin{enumerate}
	\item Test Accuracy (the predictive accuracy on unseen data)
	\item Epochs (number of passes through the entire dataset)
	\item F1 score (harmonic mean of \textit{precision} and \textit{recall}, see Eq. \ref{f1})
	\item Number of data points
	\item Precision (Positive Predictive Value, see Eq. \ref{PPV}) 
	\item Recall (True Positive Rate, see Eq. \ref{TPR})
\end{enumerate}

\begin{align}\label{f1}
F	&=	2 \cdot \dfrac{PPV\ \cdot \ TPR}{PPV\ + TPR}
\end{align}
\begin{align}\label{PPV}
PPV	&=	\dfrac{True\ Positive}{True\ Positive + False\ Positive}
\end{align}
\begin{align}\label{TPR}
TPR	&=	\dfrac{True\ Positive}{True\ Positive + False\ Negative}
\end{align}

The classification measures \textit{F1 score}, \textit{precision}, and \textit{recall} were all computed using the \texttt{classification\_report()} function of \texttt{sklearn.metrics}\cite{scikit-learn}.

\section{Results}
All experiments in this study were conducted on a laptop computer with Intel Core(TM) i5-6300HQ CPU @ 2.30GHz x 4, 16GB of DDR3 RAM, and NVIDIA GeForce GTX 960M 4GB DDR5 GPU. Table \ref{table: hyper-parameters} shows the hyper-parameters used by the DL-SVM models in the conducted experiments. Table \ref{table: summary-results} summarizes the experiment results for the presented DL-SVM models.
\begin{table}[htb!]
\centering
\caption{Hyper-parameters used in the DL-SVM models.}
		\begin{tabular}{cccc}
		\toprule
		Hyper-parameters & CNN-SVM & GRU-SVM & MLP-SVM \\
		\midrule
		Batch Size & 256 & 256 & 256 \\
		Cell Size & N/A & [256 $\times$ 5] & [512, 256, 128] \\
		No. of Hidden Layers & 2 & 5 & 3 \\
		Dropout Rate & 0.85 & 0.85 & None \\
		Epochs & 100 & 100 & 100 \\
		Learning Rate & 1e-3 & 1e-3 & 1e-3 \\
		SVM C & 10 & 10 & 0.5 \\
		\bottomrule
		\end{tabular}\\
		\label{table: hyper-parameters}
\end{table}

\begin{table}
\centering
\caption{Summary of experiment results on the DL-SVM models.}
		\begin{tabular}{cccc}
		\toprule
		Variables & CNN-SVM & GRU-SVM & MLP-SVM \\
		\midrule
		Accuracy & 77.2265625\% & 84.921875\% & 80.46875\% \\
		Data points  & 256000 & 256000 & 256000 \\
		Epochs & 100 & 100 & 100 \\
		F1 & 0.79 & 0.85 & 0.81 \\
		Precision & 0.84 & 0.85 & 0.83 \\
		Recall & 0.77 & 0.85 & 0.80 \\ 
		\bottomrule
		\end{tabular}\\
		\label{table: summary-results}
\end{table}

\indent	As opposed to what \cite{garcia2016random} did on dataset partitioning, the relative populations of each malware family were not considered in the splitting process. All the DL-SVM models were trained on $\approx$70\% of the preprocessed Malimg dataset\cite{nataraj2011malware}, i.e. 6400 malware family variants ($6400\ mod\ 256 = 0$), for 100 epochs. On the other hand, the models were tested on $\approx$30\% of the preprocessed Malimg dataset\cite{nataraj2011malware}, i.e. 2560 malware family variants ($2560\ mod\ 256 = 0$), for 100 epochs.
\begin{figure}[!htb]
	\minipage{0.45\textwidth}
	\centering
		\includegraphics[width=\linewidth]{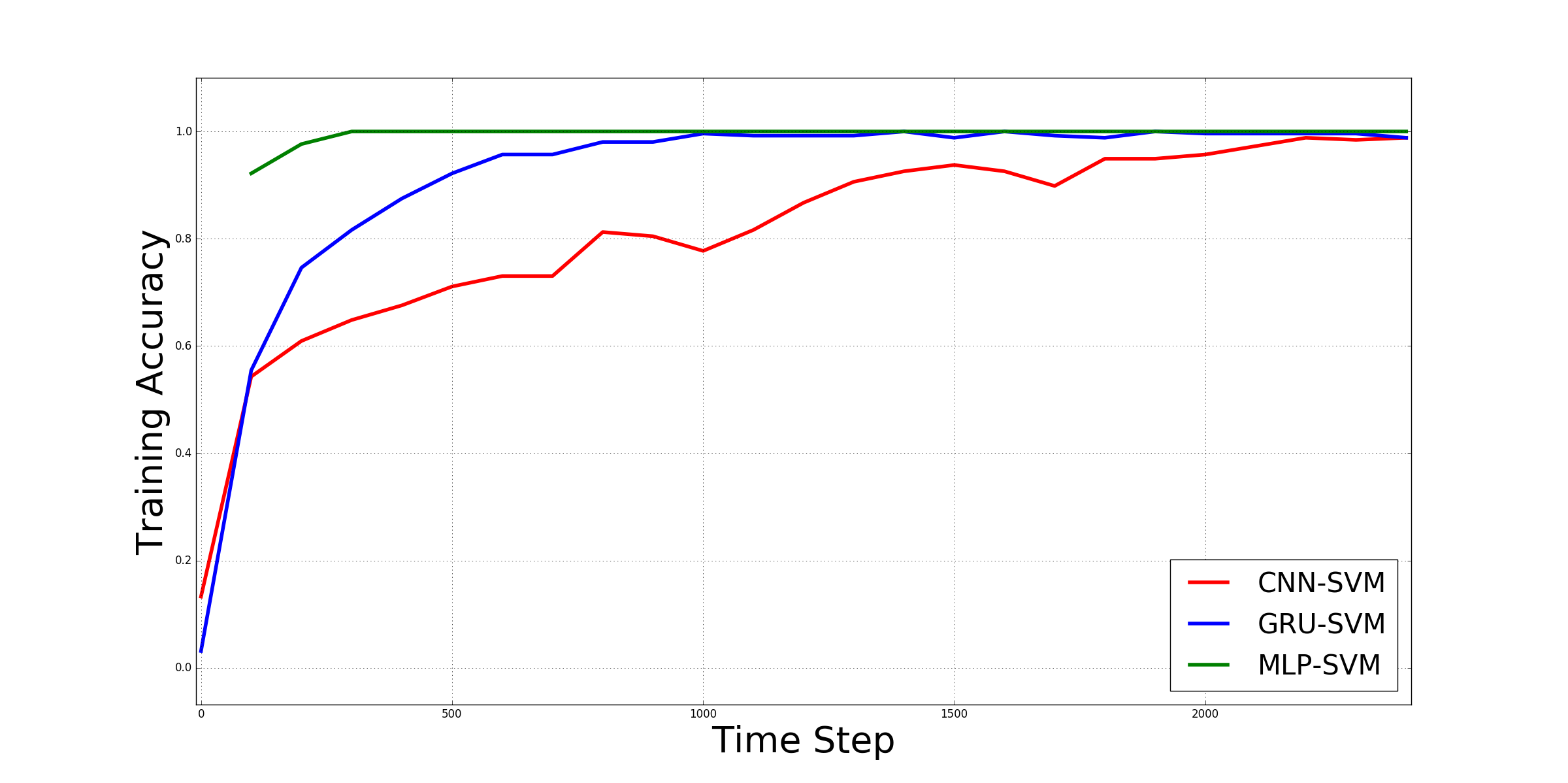}
		\caption{Plotted using \texttt{matplotlib}\cite{Hunter:2007}. Training accuracy of the DL-SVM models on malware classification using the Malimg dataset\cite{nataraj2011malware}.}
		\label{training-accuracy}
	\endminipage\hfill
\end{figure}

\indent	Figure \ref{training-accuracy} summarizes the training accuracy of the DL-SVM models for 100 epochs (equivalent to 2500 steps, since $6400 \times 100 \div 256 = 2500$). First, the CNN-SVM model accomplished its training in 3 minutes and 41 seconds with an average training accuracy of 80.96875\%. Meanwhile, the GRU-SVM model accomplished its training in 11 minutes and 32 seconds with an average training accuracy of 90.9375\%. Lastly, the MLP-SVM model accomplished its training in 12 seconds with an average training accuracy of 99.5768229\%.

\begin{figure}[htb!]
	\minipage{0.45\textwidth}
	\centering
		\includegraphics[width=\linewidth]{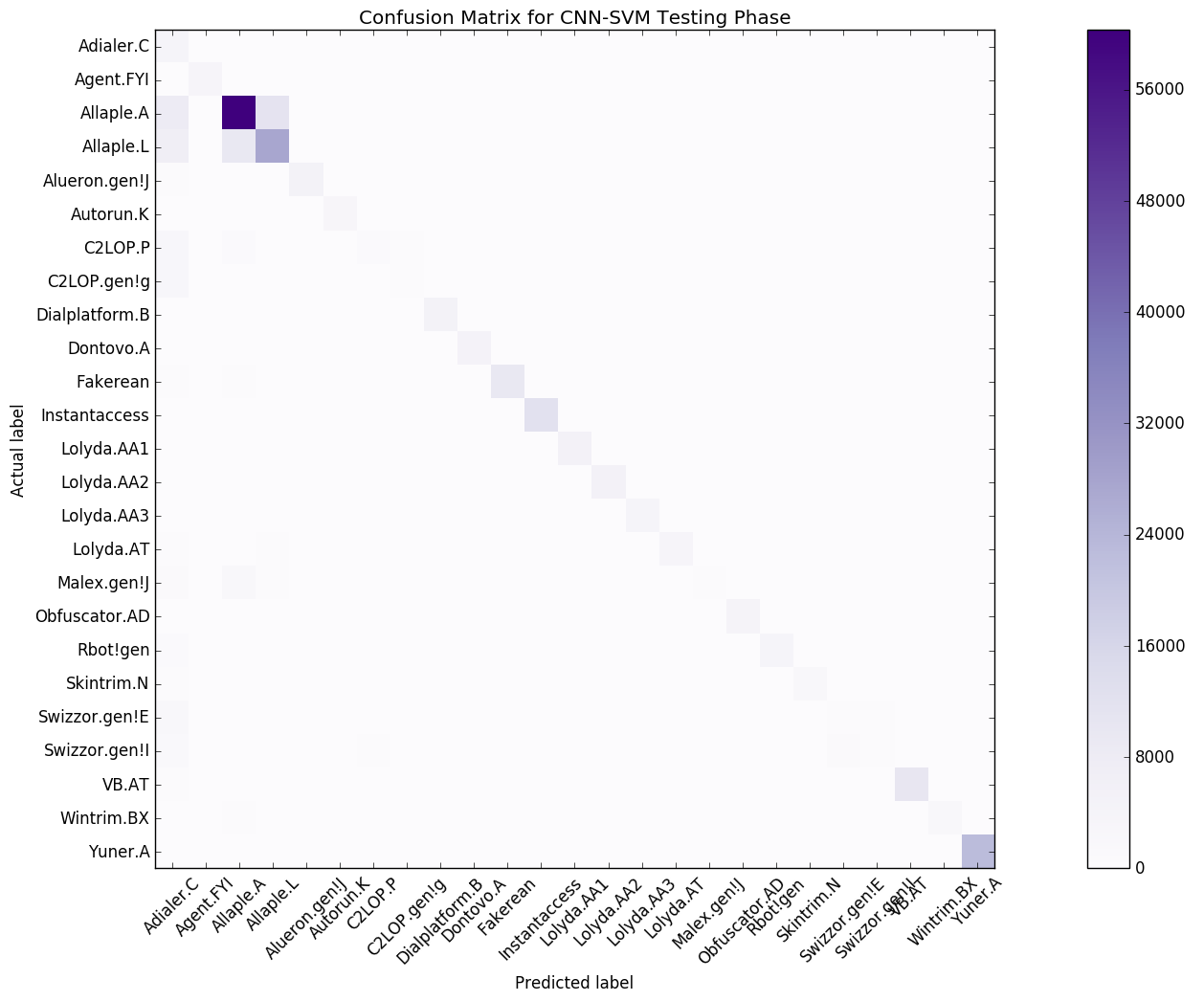}
		\caption{Plotted using \texttt{matplotlib}\cite{Hunter:2007}. Confusion Matrix for CNN-SVM testing results, showing its predictive accuracy for each malware family described in Table \ref{table: malimg-dataset}.}
		\label{conf-cnn-svm}
	\endminipage\hfill
\end{figure}

Figure \ref{conf-cnn-svm} shows the testing performance of CNN-SVM model in multinomial classification on malware families. The mentioned model had a precision of 0.84, a recall of 0.77, and a F1 score of 0.79.

\begin{figure}[!htb]
	\minipage{0.45\textwidth}
	\centering
		\includegraphics[width=\linewidth]{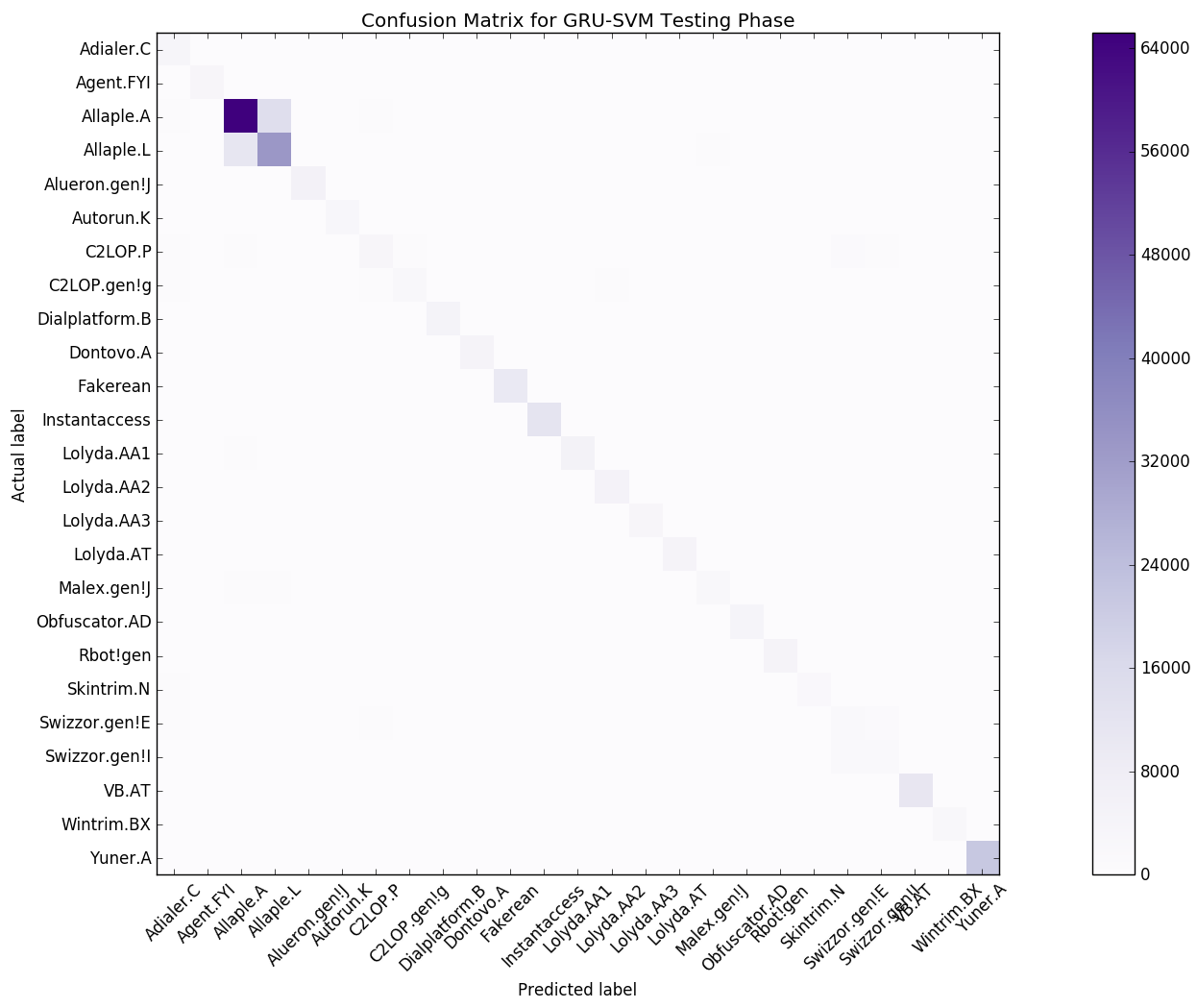}
		\caption{Plotted using \texttt{matplotlib}\cite{Hunter:2007}. Confusion Matrix for GRU-SVM testing results, showing its predictive accuracy for each malware family described in Table \ref{table: malimg-dataset}.}
		\label{conf-gru-svm}
	\endminipage\hfill
\end{figure}

Figure \ref{conf-gru-svm} shows the testing performance of GRU-SVM model in multinomial classification on malware families. The mentioned model had a precision of 0.85, a recall of 0.85, and a F1 score of 0.85.

\begin{figure}[!htb]
	\minipage{0.45\textwidth}
	\centering
		\includegraphics[width=\linewidth]{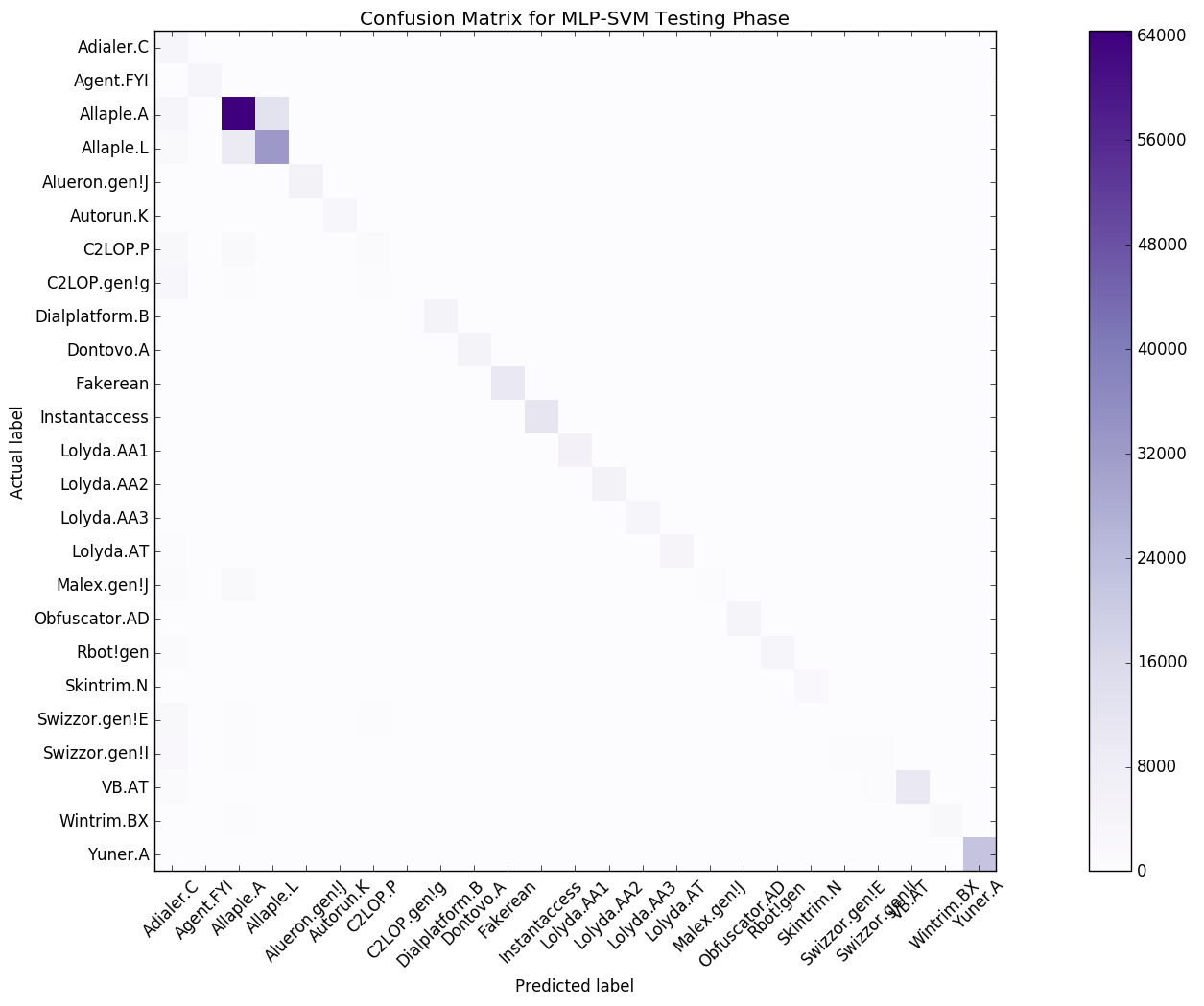}
		\caption{Plotted using \texttt{matplotlib}\cite{Hunter:2007}. Confusion Matrix for MLP-SVM testing results, showing its predictive accuracy for each malware family described in Table \ref{table: malimg-dataset}.}
		\label{conf-mlp-svm}
	\endminipage\hfill
\end{figure}

Figure \ref{conf-mlp-svm} shows the testing performance of MLP-SVM model in multinomial classification on malware families. The mentioned model had a precision of 0.83, a recall of 0.80, and a F1 score of 0.81.

As shown in the confusion matrices, the DL-SVM models had better scores for the malware families with the high number of variants, most notably, \texttt{Allaple.A} and \texttt{Allaple.L}. This may be pointed to the omission of relative populations of each malware family during the partitioning of the dataset into training data and testing data. However, unlike the results of \cite{garcia2016random}, only \texttt{Allaple.A} and \texttt{Allaple.L} had some misclassifications between them.

\section{Discussion}
It is palpable that the GRU-SVM model stands out among the DL-SVM models presented in this study. This finding comes as no surprise as the GRU-SVM model did have the relatively most sophisticated architecture design among the presented models, most notably, its 5-layer design. As explained in \cite{Goodfellow-et-al-2016}, the number of layers of a neural network is directly proportional to the complexity of a function it can represent. In other words, the performance or accuracy of a neural network is directly proportional to the number of its hidden layers. By this logic, it stands to reason that the less number of hidden layers that a neural network has, the less its performance or accuracy is. Hence, the findings in this study corroborates the literature explanation as the MLP-SVM came second (having $\approx$80.47\% test accuracy) to GRU-SVM with a 3-layer design, and the CNN-SVM came last (having $\approx$77.23\% test accuracy) with a 2-layer design.\\
\indent	The reported test accuracy of $\approx$84.92\% clearly states that the GRU-SVM model has the strongest predictive performance among the DL-SVM models in this study. This is attributed to the fact that the GRU-SVM model has the relatively most complex design among the presented models. First, its 5-layer design allows it to represent increasingly complex mappings between features and labels, i.e. function mappings $f: \vec{x} \mapsto y$. Second, its capability to learn from data of sequential nature, in which an image data belongs. This nature of the GRU-RNN comes from its gating mechanisms, given by equations in Section \ref{gru}. Through the mentioned mechanisms, the GRU-RNN solves the problem of \textit{vanishing gradients} and \textit{exploding gradients}\cite{Cho}. Thus, being able to connect information with a considerable gap. However, as indicated by the training summary given by Figure \ref{training-accuracy}, the GRU-SVM has the caveat of relatively longer computing time. Having finished its training in 11 minutes and 32 seconds, it was the slowest among the DL-SVM models. From a high-level inspection of the presented equations of each DL-SVM model (CNN-SVM in Section \ref{cnn}, GRU-SVM in Section \ref{gru}, and MLP-SVM in Section \ref{mlp}), it was a theoretical implication that the GRU-SVM would have the longest computing time as it had more non-linearities introduced in its computation. On the other hand, with the least non-linearities (having only used \texttt{LeakyReLU}), it was also theoretically implied that the MLP-SVM model would have the shortest computing time.\\
\indent	From the literature explanation\cite{Goodfellow-et-al-2016} and empirical evidence, it can be inferred that increasing the complexity of the architectural design (e.g. more hidden layers, better non-linearities) of the CNN-SVM and MLP-SVM models may catapult their predictive performance, and would be more on par with the GRU-SVM model. In turn, this implication warrants a further study and exploration that may be prolific to the information security community.

\section{Conclusion and Recommendation}
We used the Malimg dataset prepared by \cite{nataraj2011malware}, which consists of malware images for the purpose of malware family classification. We employed deep learning models with the L2-SVM as their final output layer in a multinomial classification task. The empirical data shows that the GRU-SVM model by \cite{agarap2017neural} had the highest predictive accuracy among the presented DL-SVM models, having a test accuracy of $\approx$84.92\%.\\
\indent	Improving the architecture design of the CNN-SVM model and MLP-SVM model by adding more hidden layers, adding better non-linearities, and/or using an optimized dropout, may provide better insights on their application on malware classification. Such insights may reveal an information as to which architecture may serve best in the engineering of an intelligent anti-malware system.

\section{Acknowledgment}
We extend our statement of gratitude to the open-source community, especially to TensorFlow. An appreciation as well to Lakshmanan Nataraj, S. Karthikeyan, Gregoire Jacob, and B.S. Manjunath for the Malimg dataset\cite{nataraj2011malware}.

\bibliographystyle{ACM-Reference-Format}
\bibliography{paper} 

\end{document}